\documentclass[a4paper,fleqn]{cas-dc}

\usepackage[numbers]{natbib}

\def\tsc#1{\csdef{#1}{\textsc{\lowercase{#1}}\xspace}}
\tsc{WGM}
\tsc{QE}
\tsc{EP}
\tsc{PMS}
\tsc{BEC}
\tsc{DE}
\begin{document}
\let\WriteBookmarks\relax
\def\floatpagepagefraction{1}
\def\textpagefraction{.001}
\shorttitle{Meningioma Analysis and Diagnosis using Limited Labeled Samples}
\shortauthors{Lu et~al.}

\title [mode = title]{Meningioma Analysis and Diagnosis using Limited Labeled Samples}

% ---------- Authors ----------
\author[1]{Jiamiao Lu}
\ead{241612058@sust.edu.cn}
\fnmark[1]

\author[2]{Wei Wu}
\ead{576223193@qq.com}
\fnmark[1]
\fntext[1]{These authors contributed equally.}

\author[2]{Ke Gao}
\ead{285640539@qq.com}

\author[2]{Ping Mao}
\ead{mp101010@sina.com}

\author[1]{Weichuan Zhang}
\ead{zwc2003@163.com}
\cormark[1]

\author[2]{Tuo Wang}
\ead{wt1972@tom.com}
\cormark[1]

\author[1]{Lingkun Ma}
\ead{malingkun@sust.edu.cn}
\cormark[1]

\author[4]{Jiapan Guo}
\ead{j.guo@umcg.nl}
\cormark[1]

\author[3]{Zanyi Wu}
\ead{kirby98@126.com}
\cormark[1]

\author[5]{Yuqing Hu}
\ead{ahkong23@sina.com}

\author[6]{Changming Sun}
\ead{changming.sun@csiro.au}

\affiliation[1]{organization={Shaanxi University of Science and Technology},
                city={Xi'an},
                state={Shaanxi},
                country={China}}

\affiliation[2]{
	organization={Department of Neurosurgery, The First Affiliated Hospital of Xi'an Jiaotong University},
	city={Xi'an},
	state={Shaanxi},
	postcode={710061},
	country={China}
}

\affiliation[3]{organization={Department of Neurosurgery, The First Affiliated Hospital of Fujian Medical University},
                city={Fuzhou},
                state={Fujian},
                country={China}}

\affiliation[4]{organization={Department of Radiotherapy, University of Groningen, University Medical Center Groningen},
                city={Groningen},
                country={The Netherlands}}
\affiliation[5]{
	organization={Department of Nephrology, Chenggong Hospital Affiliated to Xiamen University},
	city={Xiamen},
	state={Fujian},
	country={China}
}

\affiliation[6]{organization={CSIRO Data61},
                city={Epping},
                state={NSW},
                country={Australia}}

\cortext[cor1]{Corresponding authors.}

\begin{abstract}
The biological behavior and treatment response of meningiomas depend on their grade, making an accurate diagnosis essential for treatment planning and prognosis assessment. We observed that the weighted fusion of spatial-frequency domain features significantly influences meningioma classification performance. Notably, the contribution of specific frequency bands obtained by discrete wavelet transform varies considerably across different images. A feature fusion architecture with adaptive weights of different frequency band information and spatial domain information is proposed for few-shot meningioma learning. To verify the effectiveness of the proposed method, a new MRI dataset of meningiomas is introduced. The experimental results demonstrate the superiority of the proposed method compared with existing state-of-the-art methods in three datasets. The code will be available at: \url{https://github.com/ICL-SUST/AMSF-Net}.
\end{abstract}


\maketitle

\section{Introduction}\label{sec1}

Meningiomas, originating from arachnoid cap cells, represent the most common primary central nervous system tumor, accounting for approximately 37.6\% of all such neoplasms and exhibiting a global incidence rate of around 8.83 per 100,000 individuals, with increasing prevalence in aging populations~\cite{wang2024meningioma,price2024cbtrus}. These tumors display a wide spectrum of biological behaviors, influenced by their histological grade as defined by the World Health Organization (WHO)~\cite{Weller2024Glioma,Wang2024MolecularMeningioma,Gui2025TERTmeningioma}: benign (grade I, $\sim$80\%), atypical (grade II, $\sim$17.7\%), and malignant (grade III, $\sim$1.7\%), each associated with distinct risks of invasiveness and response to therapy~\cite{wang2024meningioma}. Globally, these tumors impose significant healthcare burdens, particularly in females and older adults, where advanced imaging modalities like magnetic resonance imaging (MRI) are essential for initial assessment~\cite{wang2024meningioma,price2024cbtrus}. MRI sequences, such as T1-weighted contrast-enhanced (T1CE), T2-weighted, and fluid-attenuated inversion recovery (FLAIR), offer insights into tumor morphology, edema, and enhancement patterns~\cite{labella2024meningioma,upreti2024meningioma}. However, histopathological examination post-resection remains the gold standard for definitive grading, supplemented by molecular markers like TERT promoter mutations or CDKN2A/B deletions in the latest WHO CNS5 updates~\cite{soni2025meningioma,sahm2025eano,groff2025tert,wach2023cdkn2a}. Despite these tools, traditional diagnostics face challenges: histopathological sampling bias fails to capture intratumoral heterogeneity, while MRI often lacks precision in predicting aggressive behavior non-invasively, due to limitations in feature validation, small sample sizes, and subjective interpretation~\cite{lucas2024spatial,patel2023radiomics,upreti2024meningioma}.

The integration of artificial intelligence (AI) has revolutionized medical imaging by enabling the automated extraction of features and recognition of patterns~\cite{dembrower2023ai,brady2024developing}. Traditional machine learning methods, including support vector machines (SVM) and random forests (RF), have been applied to radiomic features from multi-sequence MRI, achieving high area under the curve (AUC) values for distinguishing high- from low-grade meningiomas~\cite{patel2023radiomics,xiao2025mri}. Deep learning models, such as convolutional neural networks (CNNs)~\cite{Derry2023CNN} and vision transformers (ViTs)~\cite{Azad2024VisionTransformersReview,Shamshad2023TransformersSurvey}, further advance performance by learning hierarchical representations directly from images~\cite{li2023transformers,shamshad2023transformers}. Recent reviews highlight AI's role in identifying novel radiological biomarkers for meningioma detection, grading, and segmentation, with meta-analyses showing promising diagnostic accuracy in deep learning applications~\cite{maniar2023traditional,patel2023radiomics,xiao2025mri}. These advancements support personalized medicine, reducing the need for invasive procedures and optimizing clinical workflows~\cite{dembrower2023ai,brady2024developing}.
\begin{figure*}[pos=!ht]
	\centering
	\includegraphics[width=1\textwidth]{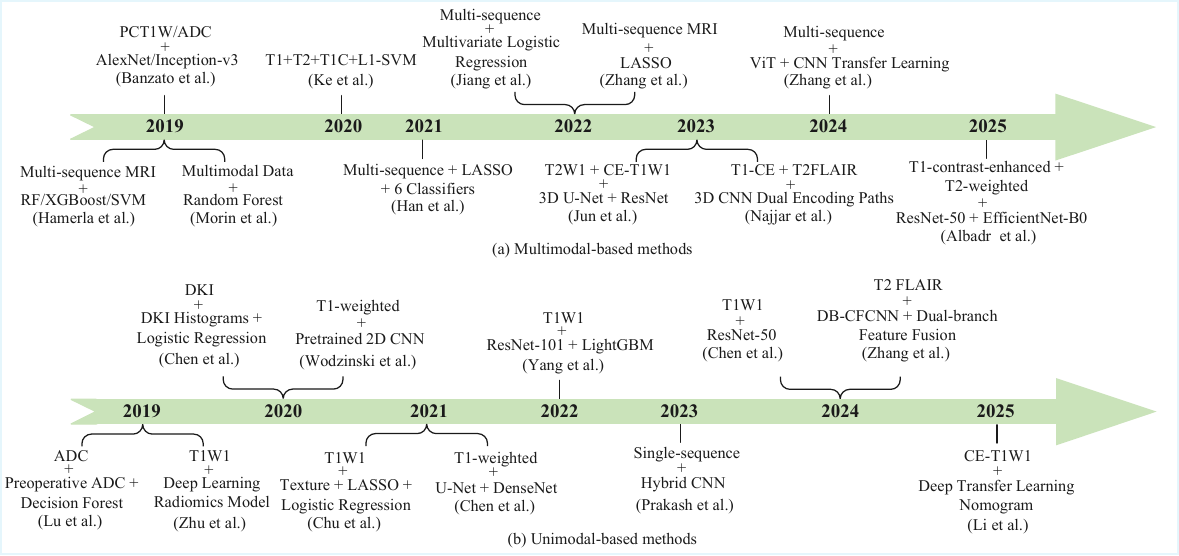}
	\caption{Timeline of the development of meningioma diagnosis and grading techniques. (a) Multimodal-based methods; (b) Unimodal-based methods.}
\label{fig1a}
\end{figure*}

Despite these progresses, key gaps remain in meningioma diagnostics~\cite{wang2024meningioma}. Histopathological grading is invasive and susceptible to sampling errors, while imaging-based approaches face data scarcity due to high MRI costs, privacy concerns, and rare subtypes, often resulting in overfitting and limited generalization~\cite{lucas2024spatial,kelly2023cybersecurity,piffer2024smalldata,labella2024multiinstitutional}. CNNs~\cite{Gao2023RFNext,Derry2023CNN} excel in local feature extraction but are limited by inductive biases that impair long-range dependency capture, essential for delineating tumor boundaries and edema~\cite{li2023transforming,azad2024vision}. Vision transformers provide global context but perform suboptimally in fine-grained local features~\cite{azad2024vision,shamshad2023transformers}. Few-shot learning has emerged to address data limitations, yet its integration with multi-resolution analysis, such as frequency-domain processing, is underexplored in meningioma diagnosis and grading~\cite{pachetti2024fewshot,patel2023radiomics,upreti2024meningioma}.

This study addresses these gaps through a comprehensive review of AI-driven methods for meningioma diagnosis and grading, emphasizing traditional machine learning and deep learning applications. We observed that the weighted fusion of spatial-frequency domain features significantly influences meningioma classification performance. Notably, the contribution of specific frequency bands obtained by the discrete wavelet transform varies considerably across different images. Then we propose an adaptive feature fusion architecture integrating spatial and frequency-domain information with a vision transformer backbone for few-shot meningioma learning. To verify the effectiveness of the proposed method, a new MRI dataset of meningiomas (i.e., XJTU Meningioma dataset) is introduced. The experimental results demonstrate the superiority of the proposed method compared with existing state-of-the-art methods in three datasets.

\section{Related Work}\label{sec2}
\enlargethispage{-\baselineskip}
In this section, existing diagnosis and grading of meningiomas methods are systematically reviewed which can be classified into two groups~\cite{wang2024meningioma,schouten2025navigating}: multimodal-based and unimodal-based methods as shown in Fig.~\ref{fig1a}. Multimodal methods integrate diverse data types, including multi-sequence MRI, imaging-derived features (e.g., radiomics and deep-learning representations), and non-imaging data (e.g., clinical and biomarker information), for enhancing diagnostic accuracy by leveraging complementary information across these domains~\cite{liu2023mmdf, zhao2024multiparametric, beuque2023combining}. In contrast, unimodal methods~\cite{upreti2024meningioma,schouten2025navigating,pahuddemortanges2024orchestrating} rely on a single imaging source, offering benefits such as simpler acquisition, less complex modeling, and reduced computational demands~\cite{luo2020aslfeat, jing2022recent,wang2024unbiased}.

%Building upon this, a further distinction is made between methods that directly classify and those that perform segmentation followed by classification. Segmentation-based approaches rely on accurate tumor boundary delineation, while classification-only methods directly learn from overall features~\cite{damiano2025impact}. Through a systematic evaluation of these methodologies, potential avenues for improvement are identified, providing new insights and directions for future research~\cite{huang2024review}.

\subsection{Multimodal-based Methods}\label{subsec2}
Multimodal-based methods aim to improve meningioma grading accuracy by integrating complementary information from multiple MRI sequences and clinical data. In the work of~\cite{morin2019integrated}, an integrated prognostic model was introduced which combines clinical variables, radiological semantic features, and radiomic features to grading prediction and uncover associations with molecular markers~\cite{lei2024semi,an2023edge,ren4962361adaptive}. To mitigate sampling bias arising from manual region of interest (ROI) selection~\cite{morin2019integrated}, Li et al.~\cite{li2019meningioma} applied 3D whole-tumor histogram analysis~\cite{Miao2025SyntheticMRIHistogram,Hwang2023SyntheticMRI_TNBC} combined with logistic regression. This method involved comprehensive volumetric assessment using conventional MRI data from 90 patients.

As foundational radiomics methods have advanced, subsequent research has emphasized advanced feature extraction and selection to boost generalizability across institutions. Ke et al.~\cite{ke2020differentiation} integrated texture and radiological features from multi-sequence MRI in 263 patients, achieving robust performance over single-sequence methods, though limited by data from only two centers. To enhance applicability, Laukamp et al.~\cite{laukamp2019accuracy} demonstrated the transferability of radiomics-based grading across heterogeneous, multi-institutional MRI datasets, supporting its feasibility in diverse scanners and protocols. Building on these integrations, multimodal strategies have increasingly targeted finer-grained tasks, such as distinguishing tumor grades and histologic subtypes. Dorfner et al.~\cite{Dorfner2025} and LaBella et al.~\cite{LaBella2024} collectively explored multiparametric MRI with SVM and RF classifiers for grade and subtype differentiation; however, small cohort sizes limited generalizability. To overcome sample size limitations, we used large cohorts and applied LASSO based feature selection followed by RF classification~\cite{ren4962361adaptive,liao2025dynamic,lu2023track,zheng2023fully}. Furthermore, Han et al.~\cite{han2021meningiomas} advanced this by benchmarking six algorithms (logistic regression~\cite{Durmaz2023RadiomicsSTEMI}, k-nearest neighbor~\cite{AlZaiti2023ECGOMI}, decision tree~\cite{Wu2023LogisticRegressionPICS}, SVM~\cite{Lin2023CTRadiomicsNodules}, RF~\cite{Shahabi2024WeightLossML}, and XGBoost~\cite{Klontzas2023PMCTRadiomics}) on multi-sequence features, highlighting SVM's discriminative edge. These efforts, however, largely centered on binary grading, leaving complex multi-class scenarios underexplored. Extending the scope, Zhang et al.~\cite{zhang2022magnetic} incorporated clinical variables, radiological semantic descriptors, and radiomic signatures to differentiate meningioma subtypes.

In parallel, multimodal MRI has proven valuable for clinical validation against standardized frameworks like the WHO classification system, which offers a surgery-oriented benchmark for grading. Jiang et al.~\cite{jiang2022efficacy} analyzed a large cohort using sequences such as T1WI, T2WI, FLAIR, DWI, and post-contrast T1, applying multivariate logistic regression to identify interpretable predictors of WHO grade and brain invasion, including indicators such as the tumor-brain interface, bone invasion, and mushroom sign that directly inform clinical decision-making. Nonetheless, manual annotation of radiological features limited quantitative depth. To refine this, Zhai et al.~\cite{zhai2021preoperative} proposed a machine learning radiomics nomogram based on multi-sequence MRI for preoperative tumor consistency classification , selecting features via variance screening and LASSO for enhanced surgical planning~\cite{wang2025principal,liao2022asrsnet,bao2022corner,lu2022image}.

The emergence of deep learning has further sophisticated multimodal integration, shifting from manual feature extraction to automated learning paradigms. Banzato et al.~\cite{Banzato2019Accuracy} pioneered the use of pre-trained convolutional neural networks on enhanced T1WI and apparent diffusion coefficient (ADC) images, illustrating transfer learning's potential for automated grading in smaller cohorts. For broader classification, Najjar et al.~\cite{Najjar2023} designed an asymmetric 3D CNN with dual encoding paths for multi-sequence T1CE and T2-FLAIR images, facilitating Grade I/II differentiation in larger cohorts, though handcrafted designs restricted adaptability. Enhancing scalability, Kaczmarczyk et al.~\cite{Kaczmarczyk2024} simplified multi-sequence frameworks, and Zhang et al.~\cite{zhang2024deep} advanced this with a meningioma feature extraction model (MFEM) combining vision transformers and CNNs to leverage transfer learning for improved representation and cross-dataset generalization~\cite{jing2021novel,islam2023background,li2023m,ma2023ct}.

Alternatively, segmentation-based strategies have also harnessed multimodal data for classification. Chen et al.~\cite{Chen2022Segmentation} utilized attention U-Net~\cite{Chen2022MeningiomaSegmentation} for automatic meningioma segmentation from multi-parametric MRI, followed by univariate screening and minimum redundancy maximum relevance (mRMR) feature selection to build an L1-regularized logistic regression model distinguishing high- from low-grade tumors. While effective, this hybrid setup constrained end-to-end learning. Advancing toward integration, Jun et al.~\cite{jun2023intelligent} introduced a two-stage model with a 3D U-Net~\cite{Gi2024DenseUNetHeadMRI} for segmentation and ResNet-based~\cite{Kang2024ResNet18PIRADS3} classification, incorporating a correlation-weighted class activation map to boost interpretability and improve decision-making in meningioma grading~\cite{gao2020fast,li2023mutual,zhang2021ndpnet,ren2024few}.

\begin{table*}[pos=!ht]
	\centering
	\setlength{\tabcolsep}{3pt}
	\footnotesize
	\linespread{1.15}\selectfont
	\caption{Multimodal-based methods.}
	\begin{tabular}{@{}>{\raggedright\arraybackslash}m{1.3cm}>{\raggedright\arraybackslash}m{3cm}>{\raggedright\arraybackslash}m{2cm}>{\raggedright\arraybackslash}m{3.5cm}>{\raggedright\arraybackslash}m{6cm}}
		\toprule
		Literatures & Data Types & \# Patients & Classification Approaches & Methods \\
		\midrule
		~\cite{morin2019integrated} & T1, T2, FLAIR, DWI, ADC & 303  &  Grade I, grade II, grade III & Radiomics feature extraction, RF, recursive partitioning analysis \\
		~\cite{li2019meningioma} & T1WI, T2WI, T1CE & 90  & Low-grade, high-grade & 3D histogram analysis, logistic regression, ROC evaluation \\
		~\cite{ke2020differentiation} & T1WI, T2WI, T1CE & 184  & Low-grade, high-grade & Radiomics feature extraction, feature selection, SVM \\
		~\cite{laukamp2019accuracy} & T1, T2, FLAIR, DWI, ADC, T1CE & 71  & Grade I, grade II & Radiomics based shape and texture analysis, feature reduction, logistic regression \\
		~\cite{Dorfner2025} & T1, T2, FLAIR, T1CE, DWI & 124  & Tumor grading & Radiomics feature extraction, SVM, RF \\
		~\cite{LaBella2024} & T1W1, T1CE, T2WI, FLAIR & 1,344  & Automated multi-label meningioma segmentation & Multi-sequence image preprocessing, deep learning, manual refinement, multi-compartment segmentation  \\
		~\cite{Banzato2019Accuracy} & PCT1W, ADC & 117  & Low-grade, high-grade & CNN \\
		~\cite{han2021meningiomas} & T1-FLAIR, T2WI, CE-T1-FLAIR & 131  & Low-grade, high-grade & LASSO, SVM, RF, XGBoost \\
		~\cite{hu2020machine} & T1WI, T2WI, T1C, ADC, SWI & 316  & Low-grade, high-grade & LASSO, RF \\
		~\cite{jiang2022efficacy} & T1WI, T2WI, FLAIR, ADC & 675 & WHO grade, brain invasion & Logistic regression \\
		~\cite{zhai2021preoperative} & T1C, T2WI, FLAIR, ADC & 172  & Tumor consistency & LASSO, logistic regression \\
		~\cite{zhang2022magnetic} & T1C, T2WI & 172  & Tumor subtype & LASSO, logistic regression \\
		~\cite{Chen2022Segmentation} & T1WI, T2WI, TICE & 143 & Low-grade, high-grade & Radiomics+machine learning \\
		~\cite{jun2023intelligent} & T1C, T2WI & 318  & Low-grade, high-grade & 3D U-Net+3D ResNet; CAM interpretability \\
		~\cite{zhang2024deep} & T1CE, T2-FLAIR & 98 & Low-grade, high-grade & CNN+vision transformer hybrid+radiomics features \\
		\bottomrule
	\end{tabular}
\end{table*}

\subsection{Unimodal-based Methods }
\label{subsec2}
Unimodal methods aim to classify meningiomas using a single MRI sequence. Compared to multimodal algorithms, unimodal methods offer advantages such as lower computational complexity, easier sample acquisition, and faster processing speed in resource-limited settings~\cite{liu2024aekan,li2019multi,wang2018survey,jing2023ecfrnet}.
\enlargethispage{-\baselineskip}
Early single-sequence approaches relied on manually engineered geometric and texture features combined with classical machine learning. Czyz et al.~\cite{czyz2017fractal} employed fractal analysis on T1WI MRI to distinguish WHO I and II meningiomas, extracting mean and maximum fractal dimensions. However, their reliance on a single feature extraction approach limited the model's ability to capture tumor complexity. To address this limitation, Hale et al.~\cite{hale2018machine} employed multiple machine learning models including KNN, SVM, and ANN with hyperparameter optimization and cross-validation, demonstrating improved classification performance. It was indicated in~\cite{friconnet2022advanced} that the multi-model based method~\cite{hale2018machine} relies on manually engineered features which may not fully characterize tumor morphology. Then a multiple geometric feature extraction strategy~\cite{friconnet2022advanced} was presented that combines fractal analysis with topological skeleton analysis to correlate features with pathological grading and brain tissue invasion for improving the classification performance~\cite{wang2020corner,zhang2023image,qiu2021recurrent,li2023traffic}.

Radiomics-based methods have been widely applied for grading meningiomas using single-sequence MRI~\cite{zhang2019corner,zhang2014corner,zhang2019discrete,zhang2020corner}. Zhang et al.~\cite{Zhang2024_physmedbiol} utilized preoperative T1CE MRI to extract texture and shape features for distinguishing high- and low-grade meningiomas, demonstrating that texture-based radiomics could effectively capture tumor heterogeneity. However, their analysis was confined to single-modality features, limiting comprehensive characterization of tumor microstructure. To enrich radiomic representation, a series of studies progressively enhanced feature diversity and robustness. Chen et al. ~\cite{chen2020histogram} introduced diffusion kurtosis imaging (DKI) based radiomics to quantify microstructural heterogeneity and identified diffusion-derived metrics (e.g., mean kurtosis and mean diffusivity) as robust discriminators of tumor grade. As pointed out in~\cite{chu2021value}, the DKI based method~\cite{chen2020histogram} relies on diffusion-only information, restricted applicability across broader MRI protocols. Then, an automatic feature extraction strategy~\cite{chu2021value} from T1WI MRI was presented for enhancing classification robustness. In the work of~\cite{duan2022radiomics}, a radiomics scoring framework was presented that integrates both radiomic and clinical parameters to improve grading consistency and interpretability across diverse patient cohorts~\cite{zhang2015contour,zhang2017noise,shui2012noise,6507646,pan2024pseudo}.

For detailed WHO classification, single-sequence MRI approaches have shown promise using traditional machine learning methods. Several studies~\cite{Bayley2022,Wang2024,Lee2025,zhu2019deep} have employed handcrafted feature extraction combined with classical classifiers. For instance, tumors from T1WI MRI were segmented and their corresponding feature representations were obtained by using a pre-trained Xception network~\cite{Sathya2024XceptionBrainTumor,Disci2025AdvancedBrainTumor}, then RF and linear discriminant analysis (LDA) were employed for classification~\cite{lu2019diagnostic}. Decision forest models have also been introduced in which clinical, morphological, and texture features from preoperative ADC images were obtained for enhancing the performance. However, these approaches rely on manual segmentation~\cite{zhu2019deep} or extensive manual feature engineering~\cite{lu2019diagnostic} which limit their clinical scalability. Building on these insights, Chen et al.~\cite{chen2019diagnostic} employed systematic feature selection methods—including distance correlation~\cite{Szekely2007DistanceCorrelation}, LASSO~\cite{Tibshirani1996Lasso} and GBDT~\cite{Friedman2001GBM}—to refine features from contrast-enhanced T1WI MRI. The combination of LASSO based feature selection and an LDA classifier demonstrated that automated feature selection enhances interpretability and consistency in three-class meningioma classification. Despite these improvements, the fundamental limitation of handcrafted features remained: they could not fully capture the complex, hierarchical patterns inherent in medical images, motivating the shift toward deep learning approaches.

To overcome the constraints of manual feature engineering, researchers began exploring deep learning methods that could automatically learn discriminative features from single-sequence MRI~\cite{Dorfner2025DeepLearningBrainTumor,Yang2023CRegKD}. In meningioma grading, most early studies employed single-modality data due to easier acquisition and standardized preprocessing~\cite{Xiao2025RadiomicsMeningiomaGrades}. Wodzinski et al.~\cite{inproceedings} classified meningiomas using preprocessed T1WI MRI that included tumor segmentation and data augmentation, applying a pre-trained 2D CNN for binary grading. However, this 2D approach could not fully capture three-dimensional tumor morphology and spatial context. To address this, Chen et al.~\cite{Chen2021Automatic} adopted a cascaded framework combining an improved U-Net for automatic segmentation and 3D convolutions for volumetric feature extraction. Using T1CE MRI with ACE sequences, their model integrated DenseNet for grading prediction, thereby enhancing spatial representation. Nevertheless, the multi-stage design increased computational cost and structural complexity. These early explorations demonstrated the potential of deep learning but also revealed the need for more streamlined and efficient architectures that could balance accuracy with computational feasibility.

Responding to the computational challenges and complexity of multi-stage pipelines, subsequent research focused on developing more elegant end-to-end architectures. Building on these advances, Yang et al.~\cite{Yang2024Deep} proposed an end-to-end SegResNet framework that simultaneously performs tumor segmentation and automatic feature extraction from segmented regions, streamlining the workflow and improving efficiency without sacrificing high accuracy in low-grade meningiomas. Direct classification approaches without explicit segmentation have also been investigated. Yang et al. ~\cite{Yang2022DLRM} proposed a deep learning radiomics model (DLRM) that extracted both radiomic and deep features from T1WI MRI using a pre-trained ResNeXt-101 network~\cite{Xie_2017_CVPR}, and employed the light gradient boosting machine (LightGBM)~\cite{Ke2017LightGBM} algorithm to construct the classifier. However, their model required the synthetic minority over-sampling technique (SMOTE)~\cite{Li2025MLHighGradeGlioma} to mitigate class imbalance, indicating potential challenges in handling imbalanced clinical datasets. To enhance model robustness, Chen et al.~\cite{Chen2024Meningioma} adopted a transfer learning approach based on ResNet-50 trained on postoperative T1WI MRI and validated it across internal and external cohorts. This approach demonstrated that transfer learning can effectively improve generalization across diverse patient populations without extensive data augmentation. With these architectural innovations established, researchers turned their attention to applying these advanced frameworks to increasingly complex and clinically relevant classification scenarios.

\begin{table*}[pos=!ht]
	\centering
	\setlength{\tabcolsep}{3pt}
	\footnotesize
	\linespread{1.3}\selectfont
	\caption{Unimodal-based methods.}
	\begin{tabular}{@{}>{\raggedright\arraybackslash}m{1.3cm}>{\raggedright\arraybackslash}m{2.8cm}>{\raggedright\arraybackslash}m{2cm}>{\raggedright\arraybackslash}m{4cm}>{\raggedright\arraybackslash}m{5.8cm}}
		\toprule
		Literatures & Data Types & \# Patients & Classification Approaches & Methods \\
		\midrule
		~\cite{czyz2017fractal} & PCT1W & 54  & WHO I, WHO II & Fractal dimension analysis , logistic regression \\
		~\cite{coroller2017radiographic} & Gadolinium-enhanced T1 & 175 & Low-grade, high-grade & Semantic and radiomic feature analysis, RF classifier \\
		~\cite{chen2020histogram} & DKI & 172 & Low-grade, high-grade & Whole tumor histogram analysis, multi-variate logistic regression \\
		~\cite{chu2021value} & T1CE& 98 & Low-grade, high-grade & Radiomics feature extraction, LASSO selection, logistic regression \\
		~\cite{duan2022radiomics} & T1CE & 151  & Low-grade, high-grade & Radiomics+machine learning models \\
		~\cite{friconnet2022advanced} & T1WI& 107  & WHO I, WHO II, WHO III & Radiomics+RF  \\
		~\cite{hale2018machine} & Conventional MRI & 128 & WHO I,WHO II & Radiomics+ANN, SVM, logistic regression \\
		~\cite{Chen2024Meningioma} & PCT1W & 1,192 & Multi-grade, pathologic markers & Fine-tuned ResNet-50 deep learning model \\
		~\cite{chen2019diagnostic} & PCT1W & 150 & Multi-grade classification & Radiomics texture + LASSO/LDA machine learning models \\
		~\cite{Bayley2022} & PCT1W & 111 & Grade I, grade II, grade III & Radiomics + SVM/RF  \\
		~\cite{Chen2021Automatic} &T1CE & 113 & Grade I, grade II, grade III & Radiomics + machine learning models\\
		~\cite{lu2019diagnostic} &T1CE & 113 & Low-grade, high-grade & Radiomics + SVM classifier \\
		~\cite{Yang2024Deep} & T1CE & 248 & Grade I, grade II, grade III & Radiomics + machine learning models \\
		~\cite{Zhang2021DeepLearningMeningioma} & T1CE & 172 & Low-grade, high-grade & Radiomics + LASSO-SVM models \\
		~\cite{Zhang2024MeningiomaConsistency} & T2-FLAIR  & 202  & Soft consistency, hard consistency& Fusion of CNN and radiomics features \\
		~\cite{Zhu2019MeningiomaCNN} & T1WI slices & Clinical dataset  & Grade I, grade II, gradeIII & Improved LeNet-5 CNN with data augmentation \\
		~\cite{Wang2024} & T1C & 164 & Grade I, grade II, grade III & Radiomics + ensemble machine learning models\\
		~\cite{inproceedings} & T1WI & 174 & Low-grade, high-grade & 2D CNN with data augmentation \\
		~\cite{yan2017potential} & T1C & 131 & Low-grade, high-grade & SVM\\
		\bottomrule
	\end{tabular}
\end{table*}

Leveraging the robust architectures developed in earlier work, recent studies have extended single-sequence approaches to tackle more challenging multi-class classification problems that better reflect clinical complexity~\cite{Ilani2025HybridDeepLearning,vanderVoort2023MultiTaskGlioma,Dorfner2025DeepLearningBrainTumor}. Beyond binary grading, Prakash et al.~\cite{Prakash2023} developed a hybrid convolutional neural network (HCNN) framework to differentiate meningioma from non-meningioma images using single-sequence T1, where key features were enhanced through ridgelet transformation and subsequently classified using CNNs architectures. Based on the classification outcomes, a segmentation algorithm was further applied to delineate tumor regions, achieving a remarkable detection accuracy of 99.81\% on the BRATS 2022 dataset. While this framework demonstrated exceptional accuracy, its sole reliance on binary tumor detection limits its direct applicability to clinical grading scenarios. Addressing the clinical demand for detailed WHO three-grade classification, several studies have explored advanced architectural designs and training strategies. Zhu et al.~\cite{Zhu2019MeningiomaCNN} employed an improved LeNet-5 convolutional neural network for three-grade WHO classification using regional MRI images. Their framework incorporated mirroring and rotation based data augmentation, as well as over-sampling techniques to address class imbalance. However, the approach relied heavily on manual annotation and architectural tuning, which may limit adaptability across varying imaging protocols. To enhance automatic feature learning, Zhang et al.~\cite{Zhang2024MeningiomaConsistency} trained a pyramid scene parsing network (PSPNet)~\cite{Zhao_2017_CVPR} to automatically detect and delineate meningiomas for three-grade classification, demonstrating that end-to-end learning can be effective in reducing dependence on manual preprocessing. While this advancement improved automation, the model primarily emphasized spatial segmentation features without adequately incorporating texture information. Recognizing that comprehensive tumor characterization requires both spatial and textural information, the same research group further proposed a dual-branch deep convolutional neural network (DB-CFCNN)~\cite{Zhang2024MeningiomaConsistency} that integrates adaptive pooling and squeeze-and-excitation modules to capture both local spatial features and fine-grained texture representations from T2-FLAIR images. By fusing deep-learned features with radiomic descriptors through multi-classifier integration, their approach achieved enhanced robustness and performance across diverse feature domains.

\begin{figure*}[pos=!ht]
	\centering
	\includegraphics[width=0.85\textwidth]{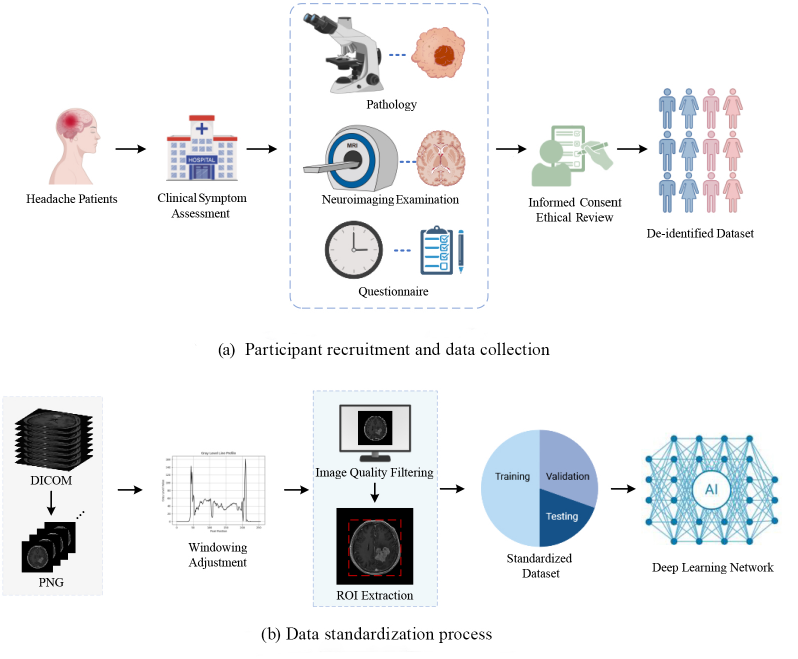}
	\caption{ Workflow of data collection and standardization. (a) Participant recruitment and clinical data acquisition. (b) MRI data preprocessing pipeline from DICOM to standardized dataset.}\label{fig1}
\end{figure*}

\section{Dataset and Proposed method}\label{sec3}
\subsection{Problem Definition}\label{subsec2}
This study focuses on few-shot meningioma learning (FML). A typical few-shot learning setting~\cite{ren2025adaptive,zhang2024re,pan2024pseudo,pan2024dycr} consists of a support set $\mathbf{S}$ and a query set $\mathbf{Q}$. The support set $\mathbf{S}$ contains $N$ distinct classes (corresponding to different pathological grades or lesion types), and each class includes only $K$ labeled samples. The query set $\mathbf{Q}$ shares the same label space as $\mathbf{S}$ but contains unlabeled samples. The goal of FML is to train a model that can accurately classify each query sample $\mathbf{q}~(\mathbf{q} \in \mathbf{Q})$ into its corresponding class using only the limited samples in the support set. Thus, the task can be formulated as an $N$-way $K$-shot meningioma learning problem.

It was indicated in~\cite{Chen2024PathFoundation,Abel2024AIMorphology,Brussee2025GNNHistopathology,Paverd2024RadiologyIntegration} that inter-class variations of medical pathology images are extremely subtle, mainly reflected in the microscopic texture distributions, nuclear morphology, and cellular structural organization. These characteristics make it challenging for the model to learn discriminative feature representations under limited data conditions~\cite{Pachetti2024FewShotReview,Huang2023SSLMedImage,Schaefer2024FoundationMultiTask}. Meanwhile, the scarcity of annotated medical data further increases the difficulty of learning transferable diagnostic representations~\cite{Huang2023SelfSupervisedReview,Lu2024VisualLanguagePathology}. To address these challenges, we adopt an episodic training strategy, in which a large number of $N$-way $K$-shot tasks are randomly constructed from the training dataset $\mathcal{D}_\text{train}$. Each task consists of a support set and a query set, continuously simulating few-shot classification scenarios during training. This enables the model to learn discriminative and transferable feature representations for FML.

\subsection{Dataset Preparation}\label{subsec2}

\subsubsection{Data Acquisition}\label{subsubsec2}
We developed a comprehensive dataset of meningioma patients based on data from the First Affiliated Hospital of Xi'an Jiaotong University. This dataset focuses on high-quality MRI scans to investigate the radiological features of meningiomas. All patients were diagnosed with meningioma, confirmed through clinical evaluation and histopathological examination according to the WHO classification of brain tumors. Only adult patients aged 18 years and older were included.

\begin{figure*}[pos=!ht]
	\centering
	\includegraphics[width=0.8\textwidth]{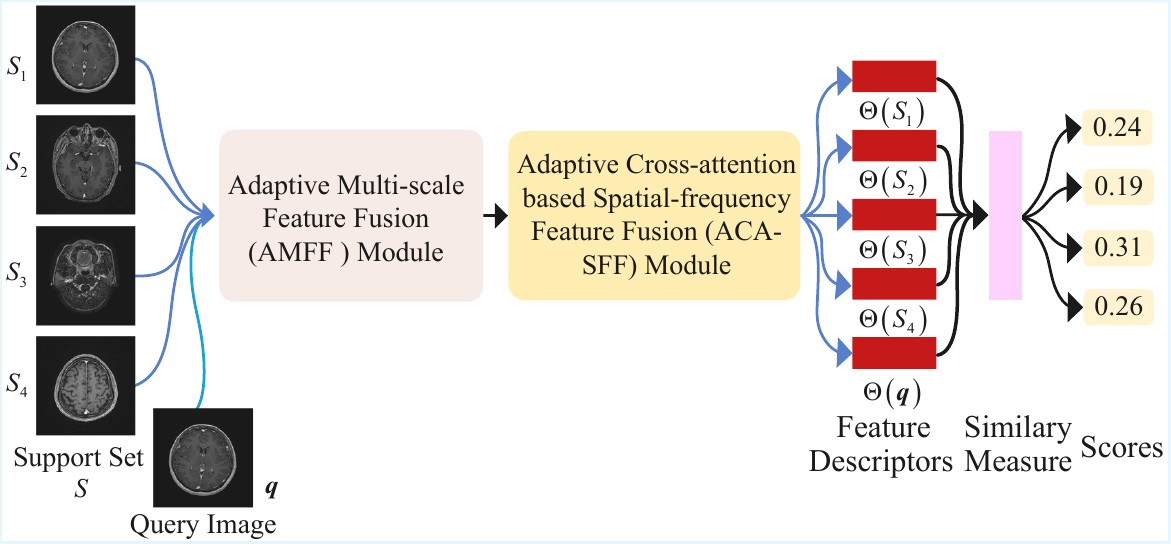}
	\caption{Overall architecture of the proposed  adaptive  multi-scale spatial-frequency network, which integrates the  adaptive multi-scale feature fusion module, the adaptive cross-attention based spatial–frequency feature fusion module, the similary module.}\label{fig3}
\end{figure*}

The dataset includes preoperative MRI scans, providing detailed anatomical information on tumor size, borders, and interactions with surrounding tissues such as edema and calcification. The high-resolution images are free from motion artifacts and other distortions, ensuring the reliability and consistency of the data for subsequent analyses.

Informed consent was obtained from all patients, and the study was approved by the Ethics Committee of the First Affiliated Hospital of Xi'an Jiaotong University, in accordance with relevant ethical guidelines. This dataset serves as a valuable resource for radiological studies of meningiomas and supports future research on tumor characterization, surgical planning, and treatment outcomes.

MRI scans were conducted for clinical purposes using a 3T Siemens Verio scanner. Acquisition parameters included the followings: repetition time (RT) is set to 1,600 ms, echo time (ET) is set to 9.4 ms, inversion time (IT) is set to 708.7 ms, flip angle is set to 150\textdegree, slice thickness 1 mm, and voxel size is set to 1 mm$\times$1 mm$\times$1 mm. Gadolinium contrast agent was administered intravenously at 0.1 mmol/kg body weight. These imaging protocols provide high-resolution, high-contrast anatomical data for detailed assessment of tumor size, boundaries, and relationships with adjacent structures. Both preoperative and postoperative MRI scans were included, ensuring a comprehensive evaluation of tumors at different stages of diagnosis and treatment. All imaging procedures adhered to the clinical protocols of Xi'an Jiaotong University First Affiliated Hospital.

\subsubsection{Ethics Statement}
This study was approved by the Ethics Committee of the First Affiliated Hospital of Xi'an Jiaotong University (approval No. Menin-202401). This is a retrospective study, written informed consent for participation was not required for this study in accordance with the national legislation and the institutional requirements. All procedures involving human participants were conducted in accordance with the ethical principles of the Declaration of Helsinki.

\subsubsection{Data Preprocessing}\label{subsubsec2}

We conducted comprehensive and rigorous preprocessing on 21 patient data provided by the First Affiliated Hospital of Xi'an Jiaotong University to ensure the quality of the data and the effectiveness of the training. It includes four categories: Class I (5 cases, 671 images), Class II (6 cases, 881 images), Class III (3 cases, 385 images), and Class N (7 cases without disease, 847 images). To ensure a balanced evaluation while preventing subject-level information leakage, the dataset was first partitioned at the patient level into training, validation, and testing sets with a 5:3:2 ratio. For each category, images were then uniformly sampled from the corresponding patient split for subsequent experiments.

In the data preprocessing stage, the original digital imaging and communications in medicine (DICOM) images are first converted into PNG format. While DICOM is widely used in medical imaging, it contains extensive metadata, and the PNG format eliminates this redundant information, making image processing and analysis more efficient and convenient. This conversion process ensures the integrity of the image data while maintaining consistency and operability for subsequent processing. To further enhance image quality, the DICOM images undergo intensity windowing, specifically adjusting the window width and window level to improve contrast and reveal finer details, thereby making the images clearer. This enhancement significantly accentuates tumor regions, improving the learning effectiveness and accuracy of subsequent models. After enhancement, a preliminary screening is conducted to remove low-quality or corrupted images, ensuring that only high-quality data suitable for analysis are retained. Following this, a cropping algorithm is applied to remove the black background around the brain tissue, focusing on retaining the primary tumor region to ensure that the model training concentrates on the relevant features, thus enhancing learning outcomes and predictive performance. Finally, a custom-developed script was used to perform patient-wise dataset partitioning, ensuring that images from the same patient never appear across different splits. This patient-disjoint protocol was consistently applied throughout training, validation, and few-shot evaluation to ensure a fair and clinically meaningful assessment.

It is worth noting that, unlike natural image benchmarks where each image can be treated as an independent instance, MRI slice data exhibit strong intra-subject correlations. Therefore, during few-shot evaluation, episodic tasks were constructed under a subject-disjoint constraint, such that support and query sets never share patients from the same patient.

\subsection{Overall Framework}\label{subsec2}

As shown in Fig.~\ref{fig3}, the proposed adaptive multi-scale spatial-frequency feature fusion network (AMSF-Net) contains two modules: an adaptive multi-scale feature fusion (AMFF) module and an adaptive cross-attention based spatial-frequency feature fusion (ACA-SFF) module. The network simultaneously leverages spatial-domain and freque-ncy-domain representations to enhance medical image classification performance. The AMFF module captures discriminative features by performing multi-level discrete wave-let decomposition and adaptively fusing the resulting multi-scale sub-bands. The ACA-SFF module enables effective spatial-frequency fusion by establishing cross-domain interactions via an adaptive cross-attention mechanism. These fused representations are then processed by a reconstruction-based episodic classifier to yield the final results.

\begin{figure*}[pos=!ht]
	\centering
	\includegraphics[width=1\textwidth]{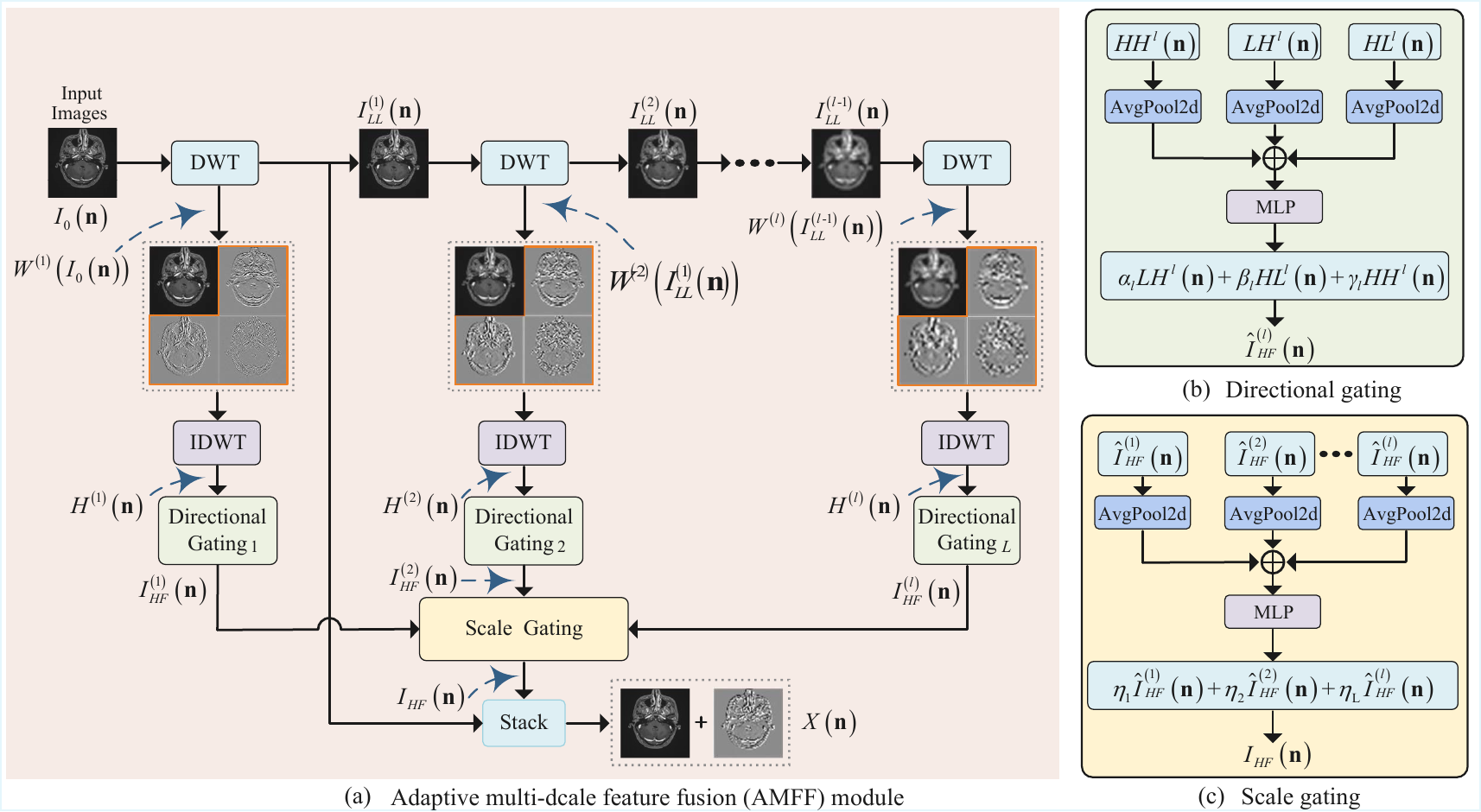}
	\caption{(a) Structure of the adaptive multi-scale feature feature fusion module.
		(b) Directional gating for adaptive fusion of \textit{LH, HL, and HH} components.
		(c) Scale gating for multi-scale high-frequency feature fusion.}\label{fig4}
\end{figure*}

\subsection{Adaptive Multi-Scale Feature Fusion (AMFF) Module}
The architecture of the designed adaptive multi-scale feature fusion (AMFF) module is shown in Fig.~\ref{fig4}(a). This module leverages hierarchical modeling and adaptive fusion to systematically extract and integrate multi-scale, multi-directional frequency information from input images, thereby delivering more discriminative frequency-aware representations for downstream networks.

For an input image $I_0(\mathbf{n})$, a cascaded strategy~\cite{ma2024crosslayer} is employed to perform an $L$-level ($l$$=$$1, 2,\ldots,L$) discrete wavelet transform (DWT). At the $l$-th ($l$$=$$1, 2,\ldots,L$) decomposition level, the low-frequency component from the previous level is used as the input, and one low-frequency subband together with three direction-specific high-frequency subbands are obtained via a Haar wavelet decomposition~\cite{kan2026haar}, which can be formulated as bellow:
\begin{equation}
	\begin{aligned}
		\left( I^{(l)}_{LL}(\mathbf{n}), I^{(l)}_{LH}(\mathbf{n}), I^{(l)}_{HL}(\mathbf{n}), I^{(l)}_{HH}(\mathbf{n}) \right)
		&= {W^{(l)}}\left( I^{(l-1)}_{LL}(\mathbf{n}) \right), \\
		I^{(0)}_{LL}(\mathbf{n}) &= I_0(\mathbf{n}),
	\end{aligned}
\end{equation}
where $I^{(l)}_{LL}(\mathbf{n})$ denotes the low-frequency component at the $l$-th ($l$$=$$1, 2,\ldots,L$) scale.  $I^{(l)}_{LL}(\mathbf{n})$ is used to characterize the global structural information of the input image. $I^{(l)}_{LH}(\mathbf{n})$, $I^{(l)}_{HL}(\mathbf{n})$, and $I^{(l)}_{HH}(\mathbf{n})$ are used to characterize the high-frequency responses along the horizontal, vertical, and diagonal directions respectively, and mainly capture local textures, edges, and fine-grained variations.

Since multi-level DWT introduces downsampling operations at each decomposition stage, high-frequency subbands at different scales exhibit inconsistent spatial resolutions.
To ensure that cross-scale and cross-directional features can be fused under a unified spatial coordinate system, the high-frequency subbands at each scale are reconstructed using the inverse discrete wavelet transform (IDWT), restoring them to the same spatial resolution as the input image~\cite{Zhang2024TMI_MAR,Wu2024TMI_WaveletSGM}.
After IDWT reconstruction, the directional high-frequency responses are denoted as $LH^{(l)}(\mathbf{n})$, $HL^{(l)}(\mathbf{n})$, and $HH^{(l)}(\mathbf{n})$.

Subsequently, the three reconstructed directional high-frequency subbands at the same scale are concatenated along the channel dimension to form a high-frequency representation as follows:
\begin{equation}
	\begin{aligned}
		H^{(l)}(\mathbf{n}) &= \mathrm{Concat}\!\left(
		LH^{(l)}(\mathbf{n}),\;
		HL^{(l)}(\mathbf{n}),\;
		HH^{(l)}(\mathbf{n})
		\right), \\
		l &= 1, 2, \ldots, L.
	\end{aligned}
\end{equation}

This design explicitly preserves multi-directional frequency information while maintaining spatial consistency, thereby providing a solid foundation for subsequent direction-adaptive fusion and cross-scale modeling.

\subsubsection{Directional Gating}

At each scale $l$ ($l$$=$$1, 2,\ldots,L$), a directional gating mechanism is introduced to adaptively fuse high-frequency responses from different directions.
Specifically, 2D average pooling (i.e., AvgPool2d) is first applied to the high-frequency features of the three directions for obtaining their corresponding directional statistical descriptors:
\begin{equation}
	\begin{aligned}
		d^{(l)}_{LH} &= \mathrm{AvgPool2d}\!\left( LH^{(l)}(\mathbf{n}) \right), \\
		d^{(l)}_{HL} &= \mathrm{AvgPool2d}\!\left( HL^{(l)}(\mathbf{n}) \right), \\
		d^{(l)}_{HH} &= \mathrm{AvgPool2d}\!\left( HH^{(l)}(\mathbf{n}) \right).
	\end{aligned}
	\label{eq:directional_descriptor}
\end{equation}

Subsequently, three directional descriptors in Equation \eqref{eq:directional_descriptor} are concatenated along the channel dimension and fed into a multi-layer perceptron (MLP). And a softmax normalization is then applied to obtain adaptive weights as follows:
\begin{equation}
	\begin{aligned}
		\left[ \alpha_l,\; \beta_l,\; \gamma_l \right]
		&= \mathrm{Softmax}\!\left(
		\mathrm{MLP}\!\left(
		d^{(l)}_{LH},\;
		d^{(l)}_{HL},\;
		d^{(l)}_{HH}
		\right)
		\right), \\
		l &= 1, 2, \ldots, L.
	\end{aligned}
\end{equation}

Based on the learned directional weights, the directionally fused high-frequency representation at the $l$-th ($l$$=$$1, 2,\ldots,L$) scale is computed as:
\begin{equation}
	\hat{I}^{(l)}_{HF}(\mathbf{n})
	= \alpha_l\, LH^{(l)}(\mathbf{n})
	+ \beta_l\, HL^{(l)}(\mathbf{n})
	+ \gamma_l\, HH^{(l)}(\mathbf{n}).
\end{equation}

As illustrated in Fig.~\ref{fig4}(b), through the proposed directional gating mechanism, the model has the capability to dynamically reweight frequency responses from different directions at each scale, thereby emphasizing directionally discriminative information in the current input while suppressing redundant or noisy directional responses.

\subsubsection{Scale Gating}

As illustrated in Fig.~\ref{fig4}(c), after completing directional fusion at each scale, a scale-gating mechanism is introduced to integrate high-frequency information across different resolution levels further. Specifically, AvgPool2d is first applied to the directionally fused high-frequency feature at each scale to obtain a scale-level descriptor:
\begin{equation}
	\begin{aligned}
		s_l &= \mathrm{AvgPool2d}\!\left( \hat{I}^{(l)}_{HF} \right), \\
		l &= 1, 2, \ldots, L.
	\end{aligned}
\end{equation}

Then, the descriptors from all scales are concatenated and fed into an MLP followed by a softmax layer to generate adaptive scale weights:
\begin{equation}
	\begin{aligned}
		\left[ \eta_1, \eta_2, \ldots, \eta_L \right]
		&= \mathrm{Softmax}\!\left(
		\mathrm{MLP}\!\left(
		s_1, s_2, \ldots, s_L
		\right)
		\right), \\
		l &= 1, 2, \ldots, L.
	\end{aligned}
\end{equation}

Finally, the multi-scale high-frequency fused representation is obtained by a weighted summation over the directionally fused high-frequency features at all scales:
\begin{equation}
	I_{HF}(\mathbf{n}) = \sum_{l=1}^{L} \eta_l \, \hat{I}^{(l)}_{HF}(\mathbf{n}).
\end{equation}

Through the proposed scale-gating mechanism, the model has the capability to adaptively select and emphasize scale-level frequency information that is more critical for the target discrimination task, according to the frequency distribution characteristics of the input sample.

At the output stage of the AMFF module, the low-frequency component obtained from the first-level decomposition, denoted as $I^{(1)}_{LL}(\mathbf{n})$, is preserved to maintain the global structural information and low-frequency background content of the input image.
Meanwhile, the fused high-frequency feature $I_{HF}(\mathbf{n})$ obtained through the proposed scale gating mechanism encodes fine-grained frequency characteristics across both scales and directions.

As illustrated in Fig.~\ref{fig4}, the final output of the AMFF module is obtained by stacking the preserved low-frequency component and the fused high-frequency feature along the channel dimension:
\begin{equation}
	X(\mathbf{n}) = \mathrm{Stack}\!\left( I^{(1)}_{LL}(\mathbf{n}),\; I_{HF}(\mathbf{n}) \right).
\end{equation}

The resulting representation $X(\mathbf{n})$ jointly incorporates global structural information and multi-scale frequency details, thereby providing a more discriminative input features for subsequent network modules.

\subsection{Adapative Cross-attention based Spatial-frequency Feature Fusion (ACA-SFF ) Module}\label{subsec2}
\begin{figure*}[pos=!ht]
	\centering
	\includegraphics[width=0.87\textwidth]{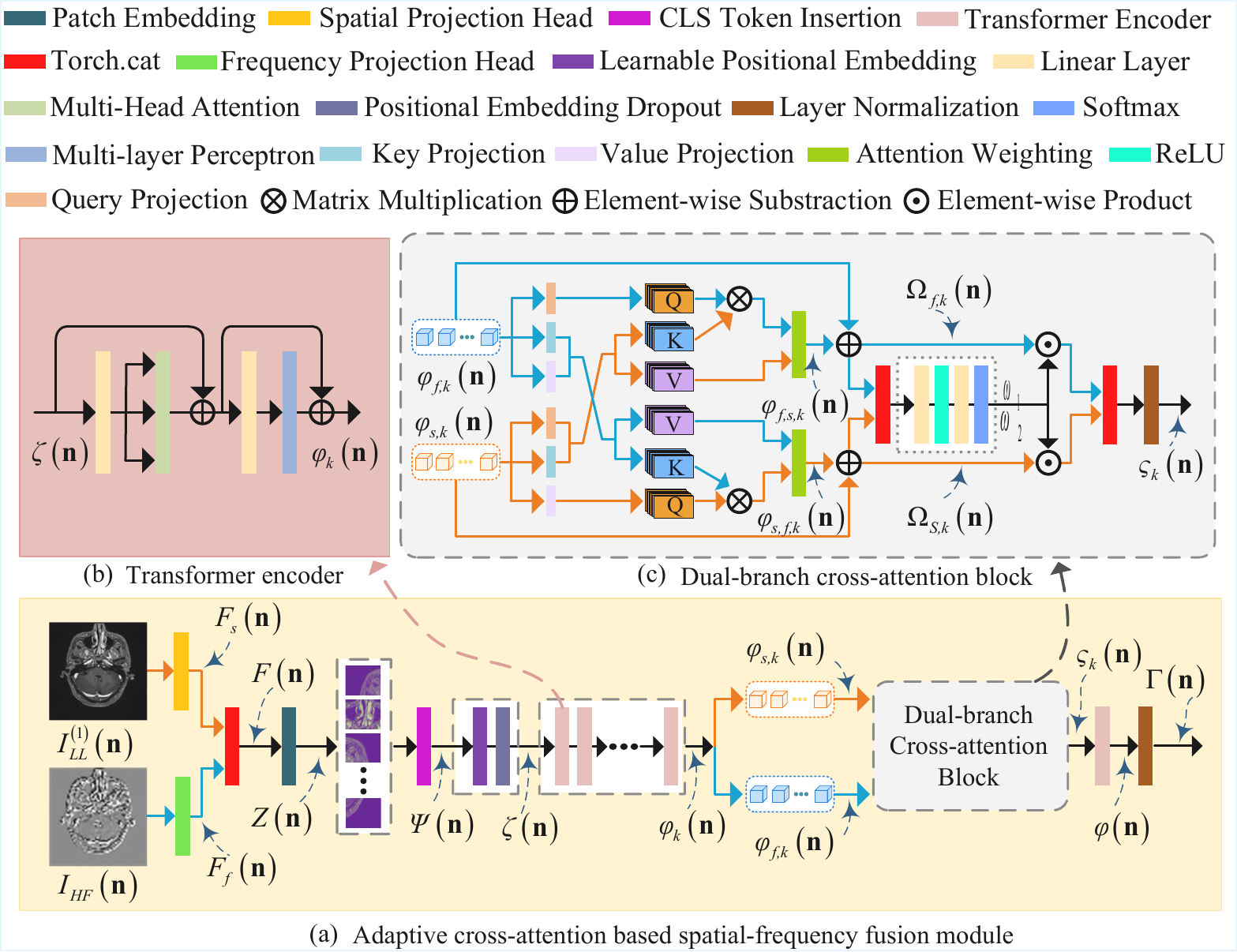}
	\caption{The architecture of the proposed ACA-SFF module which integrates dual-branch cross-attention block.}\label{fig5}
\end{figure*}

As shown in Fig.~\ref{fig5}, an adaptive cross-attention-based spatial-frequency fusion model (ACA-SFF) is presented that leverages adaptive cross-attention to achieve bidirectional spatial-frequency interaction and adaptive region-specific fusion.

Based on the output of the AMFF module, the low-frequency component $I_{LL}^{(1)}(\mathbf{n})$ is fused with the high-frequency component $I_{HF}(\mathbf{n})$ as the input of the spatial and frequency domains. Firstly, the two components are fused through their respective feature maps and then projected into the unified feature space, yielding:

\begin{equation}
	\begin{aligned}
		F_s(\mathbf{n}) &= P_s(I_{LL}^{(1)}(\mathbf{n})), \\
		F_f(\mathbf{n}) &= P_f(I_{HF}(\mathbf{n})),
	\end{aligned}
\end{equation}
where \(P_s(\cdot)\) and \(P_f(\cdot)\) represent spatial and frequency projection heads, respectively.

Later, the spatial and frequency domain features are concatenated along the channel dimension, represented as:
\begin{equation}
	F(\mathbf{n}) = \text{Concat}(F_s(\mathbf{n}), F_f(\mathbf{n})).
\end{equation}

After performing the patch embedding operation, the feature representation $F(\mathbf{n})$ is partitioned into a set of non-overlapping patches and projected into a unified feature dimension, yielding a sequence of patch-level embeddings. Subsequently, a learnable CLS token (i.e., $Z_{\mathrm{cls}}$) is prepended to the embedding sequence to form a token sequence $\Psi(\mathbf{n})$. On this basis, position encoding is added, and position embedding dropout is applied, resulting in the final token sequence $\zeta(\mathbf{n})$ that is fed into the transformer encoder, defined as follows:

\begin{equation}
	\begin{aligned}
		Z(\mathbf{n}) &= \mathrm{PatchEmbed}\!\left( F(\mathbf{n}) \right), \\
		\Psi(\mathbf{n}) &= \text{Concat}\left[ Z_{\mathrm{cls}},\, Z(\mathbf{n}) \right], \\
		\zeta(\mathbf{n}) &= \mathrm{Dropout}\!\left( \Psi(\mathbf{n}) + E_{\mathrm{pos}} \right),
	\end{aligned}
\end{equation}
where $E_{\mathrm{pos}}$ denotes the position encoding, and Dropout$(\cdot)$ represents the position embedding dropout operation. The resulting token sequence $\zeta(\mathbf{n})$ is then fed into the transformer encoder for capturing global contextual relationships and producing the encoded representation $\varphi_k(\mathbf{n})$ ($k$$=$$1, 2,\ldots,$ $K$).

As shown in Fig.~\ref{fig5}, the encoded tokens naturally preserve their domain-specific characteristics inherited from the spatial and frequency. Accordingly, $\varphi_k(\mathbf{\mathbf{n}})$ ($k$$=$$1, 2,\ldots,K$) can be regarded as consisting of spatial-domain tokens $\varphi_{s,k}(\mathbf{\mathbf{n}})$ and frequency-domain tokens $\varphi_{f,k}(\mathbf{\mathbf{n}})$. Building upon these domain-aware token representations, the ACA-SFF module performs dual-branch cross-domain attention to facilitate explicit and bidirectional information exchange between the spatial and frequency domains.

Specifically, in the cross-domain attention between the spatial and
frequency domains, the query vector $Q_s$ is obtained from the
spatial-domain token $\varphi_{s,k}(\mathbf{n})$ ($k$$=$$1, 2,\ldots,K$) through linear projection, while the key and value vectors $K_f$ and $V_f$ are obtained from the
frequency-domain token $\varphi_{f,k}(\mathbf{n})$ through linear projection,
as follows:
\begin{equation}
	\begin{aligned}
		Q_s(\mathbf{n}) &= \varphi_{s,k}(\mathbf{n}) W_Q^{(s)}, \\
		K_f(\mathbf{n}) &= \varphi_{f,k}(\mathbf{n}) W_K^{(f)}, \\
		V_f(\mathbf{n}) &= \varphi_{f,k}(\mathbf{n}) W_V^{(f)},
	\end{aligned}
\end{equation}
where $W_Q^{(s)}$, $W_K^{(f)}$, and $W_V^{(f)}$ denote learnable linear
projection matrices. The corresponding spatial-to-frequency
cross-domain attention output is expressed as:
\begin{equation}
	\begin{aligned}
	\varphi_{s,f,k}(\mathbf{n}) &= \mathrm{Attn}(Q_s(\mathbf{n}), K_f(\mathbf{n}), V_f(\mathbf{n})),\\
	 k &= 1, 2, \ldots, K.
	\end{aligned}
\end{equation}

Similarly, in the cross-domain attention from the frequency domain to
the spatial domain, the query vector $Q_f$ is obtained from the
frequency-domain token $\varphi_{f,k}(\mathbf{n})$ ($k$$=$$1, 2,\ldots,K$) through linear projection, whereas the key and value vectors $K_s$ and $V_s$ are obtained from the
spatial-domain token $\varphi_{s,k}(\mathbf{n})$ through linear projection,
given by:
\begin{equation}
	\begin{aligned}
		Q_f(\mathbf{n}) &= \varphi_{f,k}(\mathbf{n}) W_Q^{(f)}, \\
		K_s(\mathbf{n}) &= \varphi_{s,k}(\mathbf{n}) W_K^{(s)}, \\
		V_s(\mathbf{n}) &= \varphi_{s,k}(\mathbf{n}) W_V^{(s)}, \\
		k &= 1, 2, \ldots, K.
	\end{aligned}
\end{equation}

The resulting frequency-to-spatial cross-domain attention output is
formulated as:
\begin{equation}
	\begin{aligned}
	\varphi_{f,s,k}(\mathbf{n}) &= \mathrm{Attn}(Q_f(\mathbf{n}), K_s(\mathbf{n}), V_s(\mathbf{n})),\\
	k &= 1, 2, \ldots, K.
    \end{aligned}
\end{equation}

Finally, the attention operation is implemented using the scaled
dot-product formulation followed by softmax normalization:
\begin{equation}
	\mathrm{Attn}(Q, K, V) =
	\mathrm{Softmax}\!\left(\frac{QK^{\mathsf{T}}}{\sqrt{d}}\right)V,
	\label{eq:scaled_dot_attn}
\end{equation}
where $d$ denotes the feature dimension.

The outputs from the two cross-domain attention branches,
$\varphi_{f,s,k}(\mathbf{n})$ and $\varphi_{s,f,k}(\mathbf{n})$, are first combined with their
corresponding input features via residual connections, resulting in
the interaction-enhanced feature representations
$\Omega_{f,k}(\mathbf{n})$ and $\Omega_{s,k}(\mathbf{n})$, which are formulated as:
\begin{equation}
	\begin{aligned}
		\Omega_{f,k}(\mathbf{n}) &= \varphi_{f,k}(\mathbf{n}) + \varphi_{s,f,k}(\mathbf{n}), \\
		\Omega_{s,k}(\mathbf{n}) &= \varphi_{s,k}(\mathbf{n}) + \varphi_{f,s,k}(\mathbf{n}), \\
		k &= 1, 2, \ldots, K.
	\end{aligned}
\end{equation}

To achieve adaptive fusion between the frequency-domain and spatial-domain
features, the interaction-enhanced representations
$\Omega_{f,k}(\mathbf{n})$ and $\Omega_{s,k}(\mathbf{n})$ are first fed into an MLP to predict the fusion weights for the two branches.
The predicted weights are then normalized by a softmax function to
ensure non-negativity and unit sum, yielding the weight coefficients
$\omega_1$ and $\omega_2$ corresponding to
$\Omega_{f,k}(\mathbf{n})$ and $\Omega_{s,k}(\mathbf{n})$, respectively:
\begin{equation}
	\begin{aligned}
		[\omega_1, \omega_2]
		&=
		\mathrm{Softmax}\!\big(
		\mathrm{MLP}\!\left(
		[\Omega_{f,k}(\mathbf{n}),\; \Omega_{s,k}(\mathbf{n})]
		\right)
		\big), \\
		k &= 1, 2, \ldots, K.
	\end{aligned}
\end{equation}

The final fused feature is obtained via a weighted summation:
\begin{equation}
	\begin{aligned}
		\varsigma_k(\mathbf{n})
		&=
		\omega_1 \odot \Omega_{f,k}(\mathbf{n})
		+
		\omega_2 \odot \Omega_{s,k}(\mathbf{n}), \\
		k &= 1, 2, \ldots, K,
	\end{aligned}
\end{equation}
where $\odot$ denotes the Hadamard product.

Subsequently, the fused feature representation $\varsigma_k(\mathbf{n})$ ($k$$=$$1, 2,\ldots,K$) is fed
into a subsequent transformer encoder for further feature modeling.
The encoder output is expressed as:
\begin{equation}
	\varphi(\mathbf{n}) = \mathcal{T}(\varsigma_k(\mathbf{n})).
\end{equation}

The output of the transformer encoder is then normalized to obtain the
ACA-SFF output:
\begin{equation}
	\Gamma(\mathbf{n}) = \mathrm{LN}(\varphi(\mathbf{n})),
	\label{eq:layer_norm}
\end{equation}
where $\mathrm{LN}(\cdot)$ denotes the layer normalization operation.
The resulting feature representation is finally fed into
task-specific prediction heads for downstream prediction.

\subsection{Similarity Module}

A reconstruction-based metric is adopted to measure the similarity between a query image and the support classes.
The underlying assumption is that if a query sample belongs to a specific class, its feature representation can be well reconstructed from the support features of that class.

Let $\mathbf{F}_q \in \mathbb{R}^{H \times W \times d}$ denote the flattened feature map of a query image, and let $\mathcal{S}_c = \{ \mathbf{S}_{c,k} \}_{k=1}^{K}$ denote the support set of class $c$. The support features are aggregated along the shot dimension to obtain a class-level support feature representation:
\begin{equation}
	\bar{\mathbf{S}}_c = \frac{1}{K} \sum_{k=1}^{K} \mathbf{S}_{c,k}, \quad
	\bar{\mathbf{S}}_c \in \mathbb{R}^{H \times W \times d}.
\end{equation}

Both the flattened feature map of the query image $\mathbf{F}_q$ and the class-level support feature representation $\bar{\mathbf{S}}_c$ are normalized using $\ell_2$ normalization along the channel dimension.

The similarity computation is formulated as a feature reconstruction problem.
In few-shot learning scenarios, the feature dimension $d$ (e.g., 640) is typically much larger than the spatial resolution
$r = H \times W$ (e.g., 196).
Therefore, instead of explicitly constructing a projection operator, the reconstruction problem is solved in the feature space
following the dual ridge regression formulation proposed in FRN~\cite{wertheimer2021frn}. By applying the Woodbury matrix identity, the optimal closed-form reconstruction matrix is obtained as:
\begin{equation}
	\begin{aligned}
		\hat{\mathbf{W}}_c
		&=
		\left(
		\bar{\mathbf{S}}_c^{\top} \bar{\mathbf{S}}_c
		+ \lambda \mathbf{E}_d
		\right)^{-1}
		\bar{\mathbf{S}}_c^{\top} \bar{\mathbf{S}}_c, \\
		\lambda
		&=
		\mathrm{Softplus}(\alpha)\cdot\gamma + \epsilon .
	\end{aligned}
\end{equation}
where $\mathbf{E}_d \in \mathbb{R}^{d \times d}$ denotes the identity matrix,
$\alpha$ is a learnable parameter used to control the regularization strength,
and $\gamma$ and $\epsilon$ are scaling and smoothing constants, respectively.
The Softplus$(\cdot)$ function is adopted to ensure the non-negativity of the regularization coefficient $\lambda$,
thereby avoiding numerical instability.
In implementation, $\gamma$ is set to 10 and $\epsilon$ is set to 0.01.
This adaptive formulation allows the model to automatically adjust the degree of shrinkage during training.

The reconstructed query features with respect to class $c$ are then obtained
using the reconstruction matrix:
\begin{equation}
	\tilde{\mathbf{F}}_c = \mathbf{{F}}_q\,\hat{\mathbf{W}}_c .
\end{equation}

To further refine the reconstruction and compensate for potential feature misalignment,
a learnable scalar parameter $\beta$ is introduced to compute a calibration factor, and the calibrated reconstruction is obtained as:
\begin{equation}
	\begin{aligned}
		\hat{\mathbf{F}}_c
		&= \rho \cdot \tilde{\mathbf{F}}_c ,\\
		\rho
		&= 1 + \mathrm{Sigmoid}(\beta).
	\end{aligned}
\end{equation}
where $\tilde{\mathbf{F}}_c$ denotes the reconstructed query features before calibration.

Finally, the similarity between the query image and class $c$ is measured by the reconstruction quality of the query features with respect to the class-specific support subspace. Specifically, we compute the mean squared reconstruction error between each spatial token of the query feature map and its corresponding reconstructed feature, and average the errors over all spatial locations. Since a smaller reconstruction error indicates that the query sample is more consistent with the class, we take the negative mean reconstruction error as the similarity score:

\begin{equation}
	\mathrm{score}_c
	=
	-\frac{1}{H \times W}
	\sum_{i=1}^{H \times W}
	\left\|
	\mathbf{f}_{q,i}
	-
	\hat{\mathbf{f}}_{c,i}
	\right\|_2^2 ,
\end{equation}
where $\mathbf{f}_{q,i}$ denotes the $i$-th spatial token of the query
feature map $\mathbf{F}_q$, and $\hat{\mathbf{f}}_{c,i}$ denotes the
corresponding reconstructed token from $\hat{\mathbf{F}}_c$, obtained
using the support features of class $c$.

The predicted probability of the query sample belonging to class $c$ is
computed using a softmax function with a learnable temperature parameter
$\tau$:
\begin{equation}
	P(y=c \mid \mathbf{F}_q)
	=
	\frac{\exp(\tau \cdot \mathrm{score}_c)}
	{\sum_j \exp(\tau \cdot \mathrm{score}_j)} .
	\tag{28}
\end{equation}

During training, the temperature parameter $\tau$ is implemented as a learnable scaling factor,
initialized as $\tau$$=$$15.0$, and constrained within $[0.1,\,100]$ to ensure stable optimization.

\section{Experiments}\label{sec4}
\subsection{Datasets}\label{subsec2}

The proposed AMSF-Net is evaluated on three medical imaging datasets: XJTU Meningioma, Brain Tumor MRI, and COVID~\cite{Mahmoud2025MUGliomaPost,Bougourzi2023PDAttUnet}. Each dataset contains four classes, with the COVID dataset split by $70\%,15\%$ and $15\%$, and both the Brain Tumor MRI and XJTU Meningioma datasets split by $50\%,30\%$ and $20\%$ into training, validation, and test sets, respectively. The detailed class-wise sample distribution is summarized in Table~\ref{tab:multi_dataset_distribution}.
\begin{table}[!h]
\centering
\setlength{\tabcolsep}{2pt}
\caption{Sample distribution for each class and dataset.}
\label{tab:multi_dataset_distribution}
\begin{tabular}{@{}lccccc@{}}
\toprule
Dataset & Class & Train & Val & Test & Total \\
\midrule
\multirow{4}{*}{COVID}
    & COVID & 2,531 & 543 & 542 & 3,616 \\
    & Lung\_opacity & 4,208 & 902 & 902 & 6,012 \\
    & Normal & 7,134 & 1,529 & 1,529 & 10,192 \\
    & Viral pneumonia & 942 & 201 & 202 & 1,345 \\

\midrule
\multirow{4}{*}{XJTU Meningioma}
    & No\_tumor & 192 & 78 & 115 & 385 \\
    & WHO\_1 & 168 & 102 & 115 & 385 \\
    & WHO\_2 & 190 & 90 & 105 & 385 \\
    & WHO\_3 & 192 & 78 & 115 & 385 \\

\midrule
\multirow{4}{*}{Brain Tumor MRI}
    & Glioma & 810 & 486 & 324 & 1,620 \\
    & Meningioma & 817 & 491 & 327 & 1,635 \\
    & Notumor & 865 & 519 & 347 & 1,731 \\
    & Pituitary & 870 & 522 & 348 & 1,740 \\
\bottomrule
\end{tabular}
\end{table}
\begin{table*}[pos=!ht]
  \centering\tabcolsep=4pt
          \caption{Performance comparison on different datasets.}
  \label{tab:performance}
    \begin{tabular}{@{}c|c|c|cc|cc|cc@{}}
      \toprule
      \multirow{2}{*}{Train shot}
        & \multirow{2}{*}{Method}
        & \multirow{2}{*}{Backbone}
        & \multicolumn{2}{c|}{XJTU Meningioma}
        & \multicolumn{2}{c|}{Brain Tumor MRI}
        & \multicolumn{2}{c}{COVID} \\
        \cmidrule(lr){4-5}\cmidrule(lr){6-7}\cmidrule(lr){8-9}
      & & & 1-shot & 5-shot & 1-shot & 5-shot & 1-shot & 5-shot \\
      \midrule
      \multirow{9}{*}{1-shot}
        & Proto-Net~\cite{snell2017prototypical}   & ResNet-12
          & 97.79$\pm$0.04 & 98.25$\pm$0.03
          & 97.52$\pm$0.08 & 98.14$\pm$0.03
          & 89.69$\pm$0.17 & 92.81$\pm$0.06 \\
        & FRN~\cite{wertheimer2021frn}        & ResNet-12
          & 97.35$\pm$0.06 & 97.86$\pm$0.03
          & 97.32$\pm$0.08 & 98.09$\pm$0.03
          & 90.26$\pm$0.16 & 92.86$\pm$0.06 \\
        & C2-Net~\cite{ma2024c2net}     & ResNet-12
          & 97.70$\pm$0.06 & 98.64$\pm$0.06
          & 78.18$\pm$0.57 & 81.69$\pm$0.31
          & 91.10$\pm$0.34 & 93.40$\pm$0.13 \\
        & Bi-FRN~\cite{wu2023bifrn}     & ResNet-12
          & 98.50$\pm$0.08 & 97.67$\pm$0.04
          & 97.57$\pm$0.06 & 97.99$\pm$0.03
          & 90.56$\pm$0.08 & 92.87$\pm$0.14 \\
        & Proto-Net~\cite{snell2017prototypical}      & Conv-4
          & 86.95$\pm$0.14 & 94.30$\pm$0.06
          & 88.34$\pm$0.15 & 92.73$\pm$0.06
          & 77.37$\pm$0.21 & 85.86$\pm$0.09 \\
        & FRN~\cite{wertheimer2021frn}           & Conv-4
          & 78.92$\pm$0.20 & 85.22$\pm$0.09
          & 86.05$\pm$0.18 & 90.75$\pm$0.07
          & 73.92$\pm$0.23 & 82.47$\pm$0.09 \\
        & C2-Net~\cite{ma2024c2net}        & Conv-4
          & 96.65$\pm$0.23 & 97.46$\pm$0.08
          & 97.26$\pm$0.15 & 97.98$\pm$0.07
          & 89.53$\pm$0.37 & 92.23$\pm$0.14 \\
        & Bi-FRN~\cite{wu2023bifrn}        & Conv-4
          & 96.83$\pm$0.08 & 96.27$\pm$0.05
          & 97.05$\pm$0.07 & 97.70$\pm$0.04
          & 89.84$\pm$0.17 & 92.42$\pm$0.07 \\
        & Proto-Net~\cite{snell2017prototypical} & ViT-Base
          & 98.17$\pm$0.36 & 98.67$\pm$0.23
          & 95.98$\pm$0.90 & 96.57$\pm$0.41
          & 91.90$\pm$1.13 & 93.97$\pm$0.59 \\
        & \textbf{AMSF-Net (Ours)}       &ViT-Base
          & \textbf{99.70$\pm$0.01} & \textbf{99.72$\pm$0.01}
          & \textbf{99.19$\pm$0.02} & \textbf{99.33$\pm$0.02}
          & \textbf{92.51$\pm$0.15} & \textbf{94.42$\pm$0.06} \\
      \midrule
      \multirow{9}{*}{5-shot}
        & Proto-Net~\cite{snell2017prototypical}   & ResNet-12
          & 99.21$\pm$0.02 & 99.49$\pm$0.03
          & 97.57$\pm$0.07 & 98.01$\pm$0.03
          & 92.31$\pm$0.14 & 94.47$\pm$0.06 \\
        & FRN~\cite{wertheimer2021frn}        & ResNet-12
          & 99.06$\pm$0.02 & 99.33$\pm$0.03
          & 97.08$\pm$0.08 & 97.60$\pm$0.03
          & 91.51$\pm$0.14 & 93.36$\pm$0.06 \\
        & C2-Net~\cite{ma2024c2net}     & ResNet-12
          & 94.89$\pm$0.34 & 98.06$\pm$0.08
          & 97.42$\pm$0.17 & 98.10$\pm$0.07
          & 91.56$\pm$0.34 & 93.81$\pm$0.13 \\
        & Bi-FRN~\cite{wu2023bifrn}     & ResNet-12
          & 98.47$\pm$0.06 & 99.24$\pm$0.02
          & 97.99$\pm$0.06 & 98.33$\pm$0.03
          & 92.31$\pm$0.14 & 94.91$\pm$0.05 \\
        & Proto-Net~\cite{snell2017prototypical}      & Conv-4
          & 94.64$\pm$0.09 & 98.43$\pm$0.03
          & 89.61$\pm$0.14 & 95.72$\pm$0.05
          & 81.29$\pm$0.10 & 89.45$\pm$0.08 \\
        & FRN~\cite{wertheimer2021frn}          & Conv-4
          & 95.07$\pm$0.11 & 98.43$\pm$0.03
          & 92.47$\pm$0.13 & 96.46$\pm$0.05
          & 83.52$\pm$0.19 & 90.02$\pm$0.07 \\
        & C2-Net~\cite{ma2024c2net}        & Conv-4
          & 93.48$\pm$0.38 & 98.05$\pm$0.07
          & 97.56$\pm$0.13 & 98.13$\pm$0.07
          & 89.33$\pm$0.36 & 93.23$\pm$0.14 \\
        & Bi-FRN~\cite{wu2023bifrn}        & Conv-4
          & 96.85$\pm$0.09 & 98.70$\pm$0.03
          & 96.68$\pm$0.07 & 97.69$\pm$0.04
          & 88.90$\pm$0.16 & 93.54$\pm$0.06 \\
        & Proto-Net~\cite{snell2017prototypical} & ViT-Base
          & 97.87$\pm$0.45 & 99.00$\pm$0.25
          & 95.52$\pm$1.14 & 97.40$\pm$0.39
          & 88.47$\pm$1.07 & 93.85$\pm$0.59 \\
        & \textbf{AMSF-Net (Ours)}       &ViT-Base
          & \textbf{99.76$\pm$0.02} & \textbf{99.79$\pm$0.02}
          & \textbf{99.32$\pm$0.02} & \textbf{99.51$\pm$0.02}
          & \textbf{93.47$\pm$0.14} & \textbf{95.58$\pm$0.02} \\
      \bottomrule
    \end{tabular}
\end{table*}

\subsection{Implementation Details}
ViT-B/16 with ImageNet-21k pre-training~\cite{TayebiArasteh2024SelfSupervised} is adopted as the backbone for feature extraction. All experiments are conducted using the PyTorch framework on a single NVIDIA RTX 3090 GPU. The initial learning rate is set to $2\times10^{-5}$ with a weight decay of $5\times10^{-5}$. The model is trained for 8,000 epochs with the learning rate decreased by a factor of 0.5 at epochs $\{1,500,\,2,500,\,3,500,\,4,500,\,5,500\}$, and validation conducted every 500 epochs. A 300-epoch warm-up strategy is applied to stabilize training. Random cropping, horizontal flipping, rotation $(\pm10^\circ)$, and DWT transformation are applied for data augmentation. For all experiments, this paper validates the average accuracy of 10,000 randomly generated tasks on the test dataset under the standard 4-way 1-shot and 4-way 5-shot settings with a $95\%$ confidence interval.

\subsection{Performance Comparison}
In this part, we first compare the proposed AMSF-Net model with four state-of-the-art methods. The baseline methods mainly adopt ResNet-12~\cite{Feng2025EternalMAML} or Conv-4~\cite{Pachetti2024FewShotReview} as backbones; meanwhile, we additionally report the results of Proto-Net with a ViT-Base backbone as a reference under a transformer-backbone setting. ViT-B/16 is employed as the backbone to investigate transformer-based architectures for few-shot medical image analysis. The experiments encompass the XJTU Meningioma, Brain Tumor MRI,
and COVID datasets, with detailed results presented in Table~\ref{tab:multi_dataset_distribution}.

As shown in Table~\ref{tab:performance}, the proposed method consistently outperforms existing approaches on the XJTU Meningioma dataset. For instance, on our newly constructed XJTU Meningioma dataset, the proposed approach achieves superior results in both 4-way 1-shot and 4-way 5-shot classification tasks, thereby validating the efficacy of AMSF-Net. Specifically, in the 4-way 1-shot setting, AMSF-Net yields accuracy improvements of 12.75\%, 20.78\%, 3.05\%, 2.87\%, 1.91\%, 2.35\%, 2\%, and 1.2\% relative to Proto-Net~\cite{snell2017prototypical}, FRN~\cite{wertheimer2021frn}, C2-Net~\cite{ma2024c2net}, and Bi-FRN~\cite{wu2023bifrn} (with Conv-4 backbone) and the corresponding methods (with ResNet-12 backbone), respectively. In the 4-way 5-shot setting, the respective gains are 5.12\%, 4.69\%, 6.28\%, 2.91\%, 0.55\%, 0.7\%, 4.87\%, and 1.29\%.

\begin{figure}[pos=!ht]
    \centering
    \includegraphics[width=\linewidth]{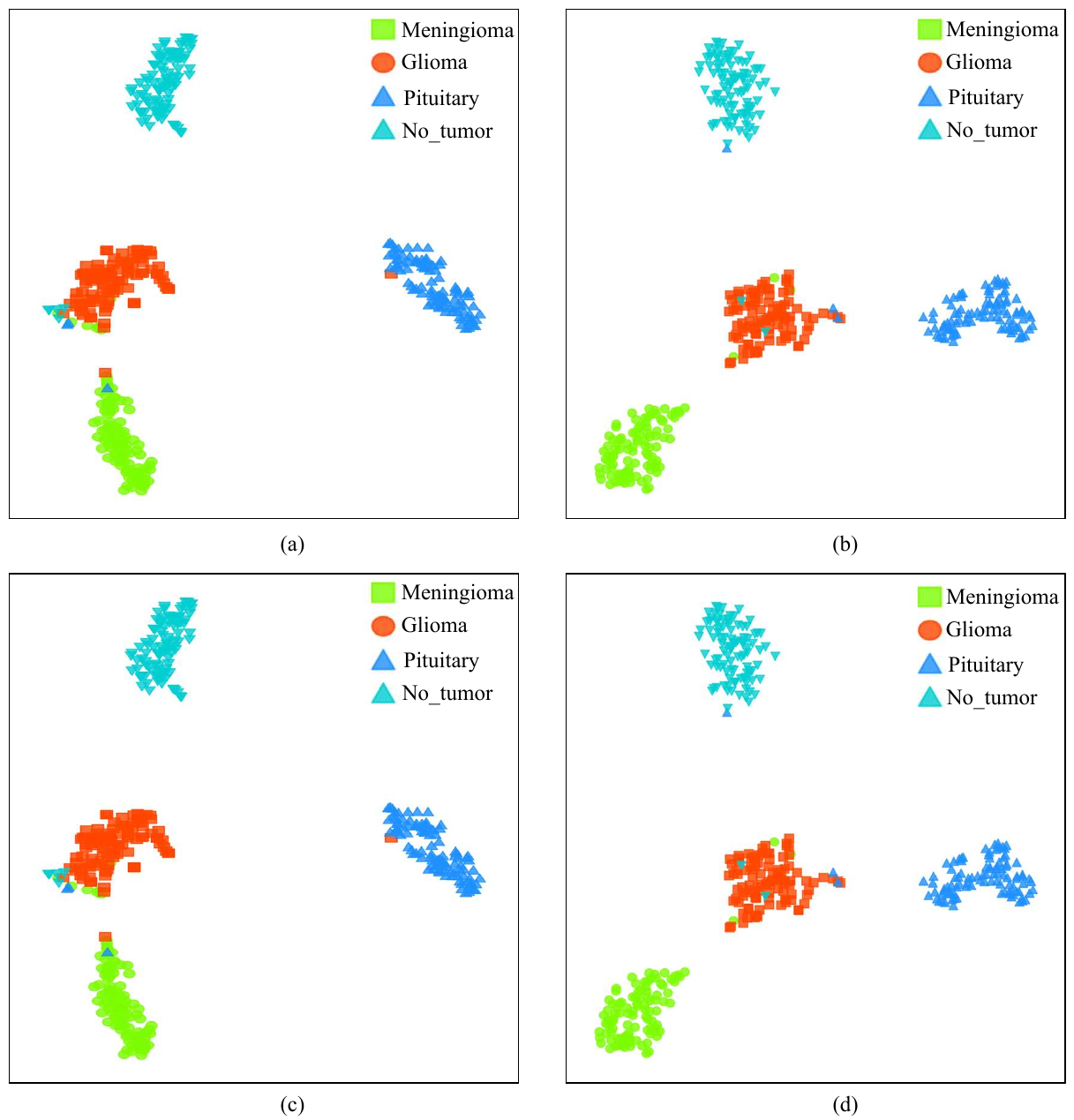}
    \caption{
        t-SNE visualization of feature distributions on the Brain Tumor MRI dataset. Subplots (a) and (b) show the feature representations learned by FRN~\cite{wertheimer2021frn} under 4-way 1-shot and 4-way 5-shot tasks, respectively; (c) and (d) show the results obtained by AMSF-Net under 4-way 1-shot and 4-way 5-shot settings, respectively.
    }
    \label{fig:tsne}
\end{figure}

To further verify the effectiveness of the proposed method, we perform t-SNE visualizations comparing AMSF-Net with the ResNet-12-based FRN~\cite{wertheimer2021frn} on the Brain Tumor MRI dataset under the 4-way 1-shot and 4-way 5-shot settings. As shown in Fig.~\ref{fig:tsne}, AMSF-Net generates more uniformly distributed and well-separated feature embeddings, demonstrating enhanced class separability.

Additionally, we perform a comparative analysis of AMSF-Net against mainstream benchmarks on the Brain Tumor MRI dataset. The findings demonstrate that AMSF-Net attains higher accuracy across all categories, particularly in challenging differentiation tasks, exhibiting superior discriminative power and underscoring the model's robustness in tumor subtype classification.

\begin{figure}[pos=!ht]
    \centering
    \includegraphics[width=\linewidth]{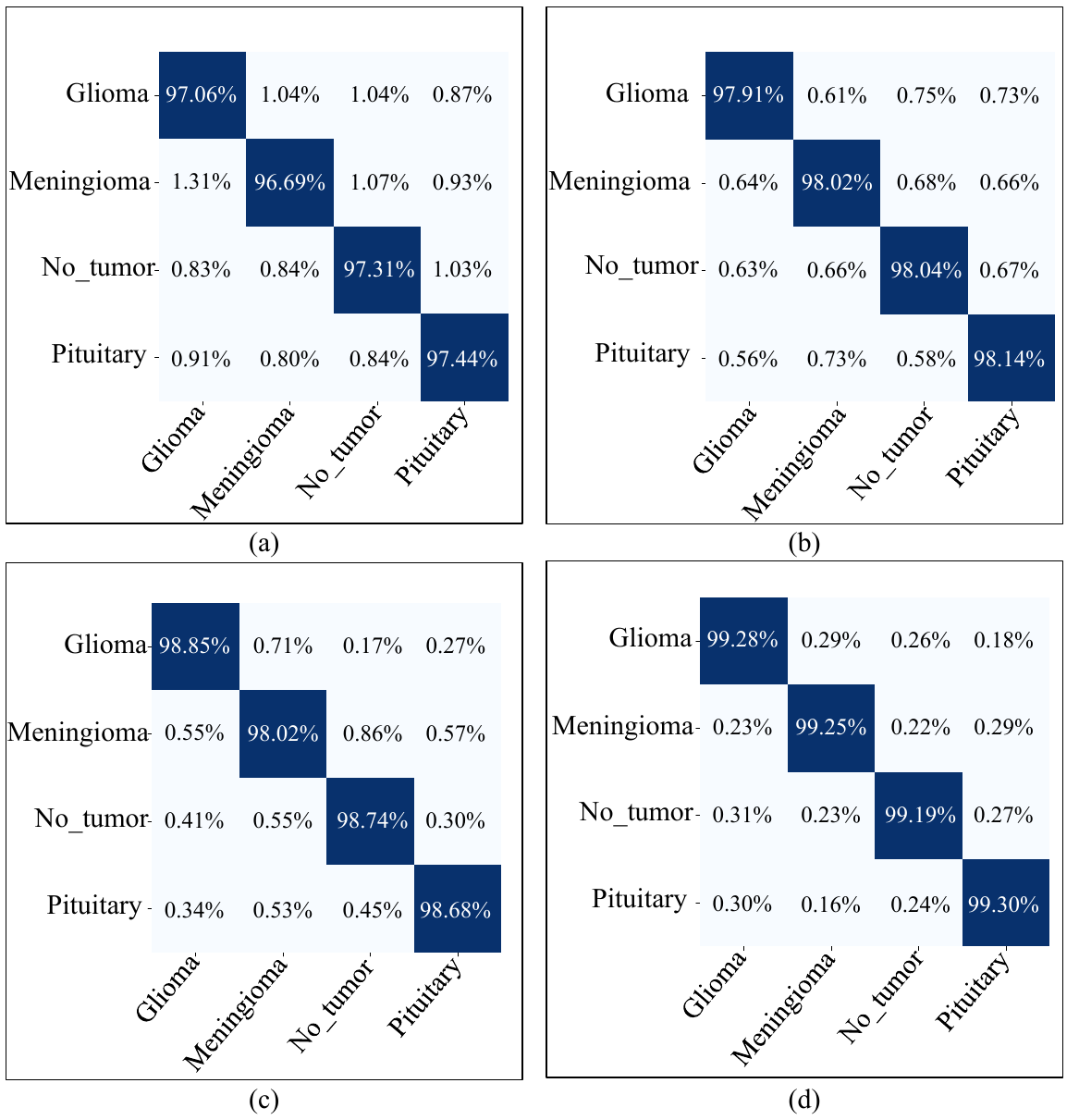}
    \caption{
         Confusion matrices on the Brain Tumor MRI dataset. (a) and (b) show the results of FRN~\cite{wertheimer2021frn} under the 4-way 1-shot and 4-way 5-shot settings, respectively; (c) and (d) show the results of AMSF-Net under the 4-way 1-shot and 4-way 5-shot settings, respectively. Each row represents the true class and each column represents the predicted class.
    }
    \label{fig:hunxiao}
\end{figure}

Additionally, as illustrated in Fig.~\ref{fig:hunxiao}, the confusion matrices on the Brain Tumor MRI dataset further validate the superiority of AMSF-Net. Compared with the ResNet-12-based FRN~\cite{wertheimer2021frn}, AMSF-Net exhibits clearer diagonal dominance and fewer off-diagonal errors under both 4-way 1-shot and 4-way 5-shot settings, indicating more accurate and stable predictions across tumor subtypes. These results demonstrate that the proposed method achieves better inter-class separability and stronger robustness in challenging tumor differentiation tasks.


\subsection{Ablation Study}

To isolate the contribution of each proposed component, we conduct an ablation study on the Brain Tumor MRI dataset under a fixed ViT-B/16 backbone with ImageNet-21k pretraining. All ablation experiments follow the same training configuration, data augmentation strategy, and evaluation protocol as described in the experimental settings.

Importantly, subject-wise data splits and subject-disjoint episodic sampling are strictly enforced throughout all experiments to ensure that no patient appears in both support and query sets, thereby preventing patient-level information leakage.

\begin{table*}[pos=!ht]
	\centering
	\caption{Ablation study on Brain Tumor MRI under the same ViT-B/16 backbone.}
	\label{tab:ablation_vit}
	\setlength{\tabcolsep}{4pt}
	\begin{tabular}{lccccc}
		\toprule
		Method (ViT-B/16) & AMFF & ACA-SFF & 1-shot & 5-shot \\
		\midrule
		ViT baseline        & \texttimes & \texttimes & 95.12$\pm$0.03& 96.22$\pm$0.01\\
		ViT + AMFF         & \checkmark & \texttimes & 96.34$\pm$0.04& 97.11$\pm$0.03\\
		ViT + ACA-SFF       & \texttimes & \checkmark & 96.77$\pm$0.02& 97.58$\pm$0.02\\
		\midrule
		AMSF-Net (Ours)     & \checkmark & \checkmark & \textbf{99.32$\pm$0.02} & \textbf{99.51$\pm$0.02} \\
		\bottomrule
	\end{tabular}
\end{table*}

\begin{table*}[pos=t]
	\centering
	\caption{Ablation study on ACA-SFF insertion depth across different datasets.}
	\label{tab:aca_sff_insertion}
	\setlength{\tabcolsep}{6pt}
	\renewcommand{\arraystretch}{1.15}
	\begin{tabular}{l c cc cc cc}
		\hline
		\multirow{2}{*}{Insertion Stage} &
		\multirow{2}{*}{Layer Index} &
		\multicolumn{2}{c}{XJTU Meningioma} &
		\multicolumn{2}{c}{Brain Tumor MRI} &
		\multicolumn{2}{c}{COVID} \\
		\cmidrule(lr){3-4}\cmidrule(lr){5-6}\cmidrule(lr){7-8}
		&  & 1-shot & 5-shot & 1-shot & 5-shot & 1-shot & 5-shot \\
		\hline
		Early Fusion   & 3  & 98.67 & 98.93 & 98.52 & 98.72 & 92.31 & 94.01 \\
		Middle Fusion  & 6  & 98.88 & 98.97 & 98.65 & 98.83 & 92.48 & 94.51 \\
		Deep Fusion    & 9  & 98.31 & 98.48 & 98.14 & 98.64 & 91.53 & 93.88 \\
		Ours & 11 &
		\textbf{99.76} & \textbf{99.79} &
		\textbf{99.32} & \textbf{99.51} &
		\textbf{93.47} & \textbf{95.58} \\
		\hline
	\end{tabular}
\end{table*}

\begin{table*}[pos=!ht]
	\centering
	\caption{Ablation study on DWT decomposition level. Results are reported as mean$\pm$std over 3 runs. Best results are in bold.}
	\label{tab:ablation_dwt_vit}
	\setlength{\tabcolsep}{3pt}
	\renewcommand{\arraystretch}{1.1}
	\begin{tabular}{l c cc cc cc c}
		\toprule
		\multirow{2}{*}{Model} & \multirow{2}{*}{Backbone} &
		\multicolumn{2}{c}{XJTU Meningioma} &
		\multicolumn{2}{c}{Brain Tumor MRI} &
		\multicolumn{2}{c}{COVID}  \\
		\cmidrule(lr){3-4}\cmidrule(lr){5-6}\cmidrule(lr){7-8}
		& & 1-shot & 5-shot & 1-shot & 5-shot & 1-shot & 5-shot & \\
		\midrule
		AMSF-Net (Ours) (DWT level=1) & ViT &  99.32$\pm$0.02&  99.35$\pm$0.02&  98.82$\pm$0.04&  98.96$\pm$0.03&  92.81$\pm$0.14&  94.89$\pm$0.05&  \\
		AMSF-Net (Ours) (DWT level=2) & ViT &  99.46$\pm$0.01&  99.47$\pm$0.01&  99.02$\pm$0.04&  99.26$\pm$0.02&  93.02$\pm$0.14&  94.93$\pm$0.05&  \\
		AMSF-Net (Ours) (DWT level=3) & ViT & \textbf{99.76$\pm$0.02} &  \textbf{99.79$\pm$0.02}&  \textbf{99.32$\pm$0.02}&  \textbf{99.51$\pm$0.02}&  \textbf{93.47$\pm$0.14}&  \textbf{95.58$\pm$0.02}  \\
		AMSF-Net (Ours) (DWT level=4) & ViT & 99.55$\pm$0.02 &  99.56$\pm$0.02&  99.14$\pm$0.02&  99.27$\pm$0.02&  93.08$\pm$0.13&  95.04$\pm$0.03&  \\
		\bottomrule
	\end{tabular}
\end{table*}
\subsubsection{The Impact of AMFF and ACA-SFF on Performance}

We conduct an ablation study to investigate the contribution of each proposed component, including the AMFF module and the ACA-SFF module.

As summarized in Table~\ref{tab:ablation_vit}, introducing AMFF into the vanilla ViT baseline leads to a noticeable performance improvement. Specifically, the 1-shot accuracy increases from 95.12\% to 96.34\%, while the 5-shot accuracy improves from 96.22\% to 97.11\%. This result highlights the effectiveness of explicitly modeling multi-scale frequency-domain information.

Building upon AMFF, the inclusion of the ACA-SFF module further boosts performance by enabling bidirectional interactions between spatial and frequency representations. With ACA-SFF, the model achieves 96.77\% accuracy in the 1-shot setting and 97.58\% in the 5-shot setting.

When both modules are jointly integrated, the complete AMSF-Net delivers the best performance across all evaluation settings, reaching 99.32\% accuracy for 1-shot classification and 99.51\% for 5-shot classification. Notably, the combined improvement consistently surpasses that obtained by either module alone, indicating a strong complementarity between AMFF and ACA-SFF. While AMFF enriches multi-scale frequency representations, ACA-SFF adaptively fuses spatial and frequency features through cross-attention.

Overall, the ablation results demonstrate that the performance gains are not solely attributable to the ViT backbone or large-scale pretraining, but arise from the proposed spatial-frequency fusion mechanisms.

\subsubsection{The Impact of ACA-SFF Insertion Layer}

We study the effect of ACA-SFF insertion depth by placing it at different backbone stages while
keeping all other settings unchanged. As shown in Table~\ref{tab:aca_sff_insertion}, we compare early fusion (layer index 3),
middle fusion (layer index 6), and deep fusion (layer index 9). The results indicate that the
insertion depth noticeably affects few-shot performance across datasets. In particular, inserting
ACA-SFF at a deeper layer achieves the best overall performance, and thus, we adopt layer index 11
as the default setting in subsequent experiments.

\subsubsection{The Impact of DWT Decomposition Level}

Table~\ref{tab:ablation_dwt_vit} presents an ablation study on the number of DWT decomposition levels $L$. As $L$ increases from 1 to 3, the classification performance consistently improves across all datasets and both 1-shot and 5-shot settings, indicating that multi-scale frequency representations are beneficial for few-shot medical image classification. Shallow decomposition ($L$$=$$1$) captures limited frequency information and fails to adequately model fine-grained texture patterns, while moderate depths ($L$$=$$2,3$) effectively balance global structural cues and localized high-frequency details. However, further increasing the decomposition depth to $L$$=$$4$ leads to performance saturation or slight degradation, which can be attributed to excessive frequency fragmentation, noise amplification, and information loss introduced by repeated downsampling and reconstruction. Overall, $L$$=$$3$ achieves the best trade-off between representation richness and robustness, and is therefore adopted as the default setting.

\section{Conclusion}\label{sec5}
This study proposes an innovative AMSF-Net framework, which achieves significant advancements in the challenging task of meningioma grading and classification under limited annotated data. By explicitly integrating spatial and frequency domain features and leveraging a domain-aware transformer architecture, the proposed approach substantially enhances the model’s feature representation and discriminative ability in few-shot learning scenarios. An adaptive fusion mechanism further strengthens the extraction of critical imaging features, enabling the model to consistently outperform existing state-of-the-art methods in meningioma classification tasks. Extensive experiments conducted on both our self-constructed and soon-to-be-released XJTU meningioma dataset, as well as other public benchmark datasets, comprehensively validate the effectiveness and robustness of the proposed method. These findings demonstrate its considerable potential and academic value for intelligent diagnosis of meningioma.

\bibliographystyle{cas-model2-names}
\bibliography{cas-refs}
\end{document}